\pdfoutput=1

\documentclass[11pt]{article}

\usepackage{EMNLP2022}

\usepackage{times}
\usepackage{latexsym}
\usepackage{amsmath}
\usepackage{amssymb}
\usepackage{mathtools}
\usepackage{comment}
\usepackage{graphicx}
\usepackage{caption}
\usepackage{subcaption}
\usepackage{enumitem}
\usepackage{bbm}

\usepackage[T1]{fontenc}

\usepackage[utf8]{inputenc}

\usepackage{microtype}

\usepackage{inconsolata}

\title{Dictionary-Assisted Supervised Contrastive Learning}

\author{Patrick Y. Wu$^1$, Richard Bonneau$^{1,2,4,5}$, Joshua A. Tucker$^{1,2,3}$, \and Jonathan Nagler$^{1,2,3}$ \\
  $^1$ Center for Social Media and Politics, New York University \\
  $^2$ Center for Data Science, New York University \\
  $^3$ Department of Politics, New York University\\
  $^4$ Department of Biology, New York University\\
  $^5$ Courant Institute of Mathematical Sciences, New York University \\
  \texttt{\{pyw230, bonneau, joshua.tucker, jonathan.nagler\}@nyu.edu} \\}

\begin{document}
\maketitle

\begin{abstract}
Text analysis in the social sciences often involves using specialized dictionaries to reason with abstract concepts, such as perceptions about the economy or abuse on social media. These dictionaries allow researchers to impart domain knowledge and note subtle usages of words relating to a concept(s) of interest. We introduce the dictionary-assisted supervised contrastive learning (DASCL) objective, allowing researchers to leverage specialized dictionaries when fine-tuning pretrained language models. The text is first keyword simplified: a common, fixed token replaces any word in the corpus that appears in the dictionary(ies) relevant to the concept of interest. During fine-tuning, a supervised contrastive objective draws closer the embeddings of the original and keyword-simplified texts of the same class while pushing further apart the embeddings of different classes. The keyword-simplified texts of the same class are more textually similar than their original text counterparts, which additionally draws the embeddings of the same class closer together. Combining DASCL and cross-entropy improves classification performance metrics in few-shot learning settings and social science applications compared to using cross-entropy alone and alternative contrastive and data augmentation methods.\footnote{Our code is available at \url{https://github.com/SMAPPNYU/DASCL}.}
\end{abstract}

\section{Introduction}
We propose a supervised contrastive learning approach that allows researchers to incorporate dictionaries of words related to a concept of interest when fine-tuning pretrained language models. It is conceptually simple, requires low computational resources, and is usable with most pretrained language models.

Dictionaries contain words that hint at the sentiment, stance, or perception of a document \citep[see, e.g.,][]{fei-etal-2012-dictionary}. Social science experts often craft these dictionaries, making them useful when the underlying concept of interest is abstract \citep[see, e.g.,][]{brady_wills_jost_tucker_bavel_2017,young_soroka_2012}.  Dictionaries are also useful when specific words that are pivotal to determining the classification of a document may not exist in the training data. This is a particularly salient issue with small corpora, which is often the case in the social sciences. 

\begin{figure}
    \centering
    \includegraphics[width=0.49\textwidth]{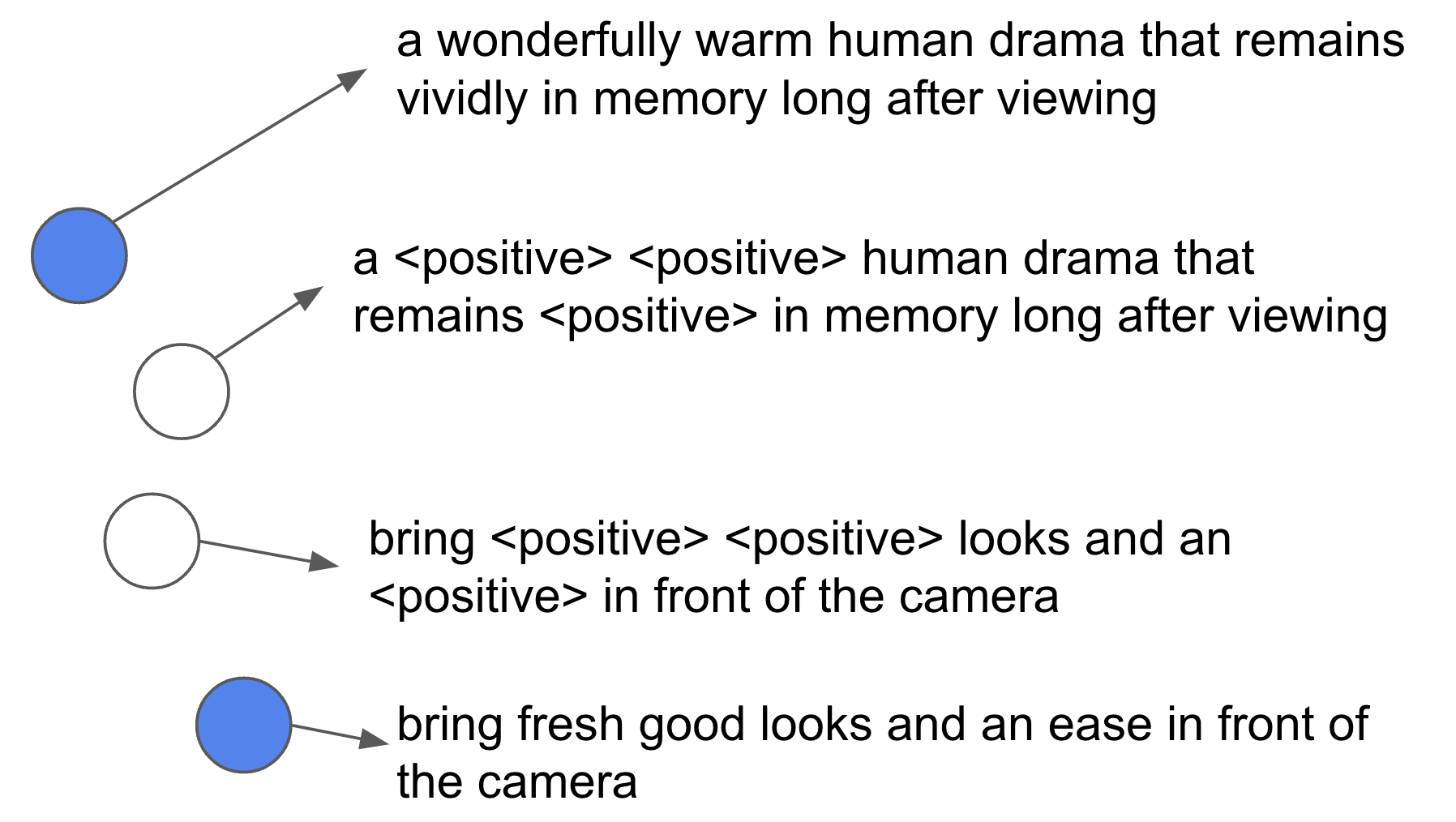}
    \caption{The blue dots are embeddings of the original reviews and the white dots are the embeddings of the keyword-simplified reviews from the SST-2 dataset \citep{wang-etal-2018-glue}. Both reviews are positive, although they do not overlap in any positive words used. The reviews are more textually similar after keyword simplification. Using \textsc{BERT$_{\text{base-uncased}}$} out-of-the-box, the cosine similarity between the original reviews is $.654$ and the cosine similarity between the keyword-simplified reviews is $.842$. Although there are some issues with using cosine similarity with BERT embeddings \citep[see, e.g., ][]{ethayarajh-2019-contextual,zhou-etal-2022-problems}, we use it as a rough heuristic here.}
    \label{fig:dascl_intuition}
\end{figure}

However, recent supervised machine learning approaches do not use these dictionaries. We propose a contrastive learning approach, dictionary-assisted supervised contrastive learning (DASCL), that allows researchers to leverage these expert-crafted dictionaries when fine-tuning pretrained language models. We replace all the words in the corpus that belong to a specific lexicon with a fixed, common token. When using an appropriate dictionary, keyword simplification increases the textual similarity of documents in the same class. We then use a supervised contrastive objective to draw together text embeddings of the same class and push further apart the text embeddings of different classes \citep{khosla_teterwak_wang_tian_isola_maschinot_liu_2020,gunel2020supervisedcontrastive}. Figure \ref{fig:dascl_intuition} visualizes the intuition of our proposed method. 

The contributions of this project are as follows.
\begin{itemize}[noitemsep]
    \item We propose keyword simplification, detailed in Section \ref{sec:keyword_simplification}, to make documents of the same class more textually similar. 
    \item We outline a supervised contrastive loss function, described in Section \ref{sec:dascl_objective_details}, that learns patterns within and across the original and keyword-simplified texts. 
    \item We find classification performance improvements in few-shot learning settings and social science applications compared to two strong baselines: (1) \textsc{RoBERTa} \citep{liu2019roberta} / \textsc{BERT} \citep{devlin-etal-2019-bert} fine-tuned with cross-entropy loss, and (2) the supervised contrastive learning approach detailed in \citet{gunel2020supervisedcontrastive}, the most closely related approach to DASCL. To be clear, although \textsc{BERT} and \textsc{RoBERTa} are not state-of-the-art pretrained language models, DASCL can augment the loss functions of state-of-the-art pretrained language models.
\end{itemize}

\section{Related Work} 
\textbf{Use of Pretrained Language Models in the Social Sciences}. Transformers-based pretrained language models have become the de facto approach when classifying text data \citep[see, e.g., ][]{devlin-etal-2019-bert,liu2019roberta,raffelT52020}, and are seeing wider adoption in the social sciences. \citet{terechshenko_etal_2021} show that RoBERTa and XLNet \citep{yang_dai_yang_carbonell_salakhutdinov_russ_le_2019_xlnet} outperform bag-of-words approaches for political science text classification tasks. \citet{ballard_detamble_dorsey_heseltine_johnson_2022} use BERTweet \citep{nguyen-etal-2020-bertweet} to classify tweets expressing polarizing rhetoric. \citet{lai_etal_2022} use BERT to classify the political ideologies of YouTube videos using text video metadata. DASCL can be used with most pretrained language models, so it can potentially improve results across a range of social science research. 

\textbf{Usage of Dictionaries}. Dictionaries play an important role in understanding the meaning behind text in the social sciences. \citet{brady_wills_jost_tucker_bavel_2017} use a moral and emotional dictionary to predict whether tweets using these types of terms increase their diffusion within and between ideological groups. \citet{simchon_brady_vanbavel_2022_trollanddivide} create a dictionary of politically polarized language and analyze how trolls use this language on social media. \citet{hopkins_kim_kim_2017_newspapercoverage} use dictionaries of positive and negative economic terms to understand perceptions of the economy in newspaper articles. Although dictionary-based classification has fallen out of favor, dictionaries still contain valuable information about usages of specific or subtle language. 

\textbf{Text Data Augmentation}. Text data augmentation techniques include backtranslation \citep{sennrich-etal-2016-neural} and rule-based data augmentations such as random synonym replacements, random insertions, random swaps, and random deletions \citep{wei-zou-2019-eda,karimi-etal-2021-aeda-easier}. \citet{shorten_etal_2021_textdataaugmentation} survey text data augmentation techniques. \citet{longpre-etal-2020-effective} find that task-agnostic data augmentations typically do not improve the classification performance of pretrained language models. We choose dictionaries for keyword simplification based on the concept of interest underlying the classification task and use the keyword-simplified text with a contrastive loss function. 

\textbf{Contrastive Learning}. Most works on contrastive learning have focused on self-supervised contrastive learning. In computer vision, images and their augmentations are treated as positives and other images as negatives. Recent contrastive learning approaches match or outperform their supervised pretrained image model counterparts, often using a small fraction of available annotated data \citep[see, e.g.,][]{pmlr-v119-chen20j,He_2020_CVPR,chen_fan_girshick_2020_mocov2,grill_etal_2020_byol}. Self-supervised contrastive learning has also been used in natural language processing, matching or outperforming pretrained language models on benchmark tasks \citep[see, e.g.,][]{fang_etal_2020_cert,klein-nabi-2020-contrastive}. 

Our approach is most closely related to works on supervised contrastive learning. \citet{wen_etal_2016_centerloss} propose a loss function called center loss that minimizes the intraclass distances of the convolutional neural network features. \citet{khosla_teterwak_wang_tian_isola_maschinot_liu_2020} develop a supervised loss function that generalizes NT-Xent \citep{pmlr-v119-chen20j} to an arbitrary number of positives. Our work is closest to that of \citet{gunel2020supervisedcontrastive}, who also use a version of NT-Xent extended to an arbitrary number of positives with pretrained language models. Their supervised contrastive loss function is detailed in Section \ref{sec:gunelscllossfunction}.

\section{Method}
\begin{figure}
    \centering
    \includegraphics[width=0.48\textwidth]{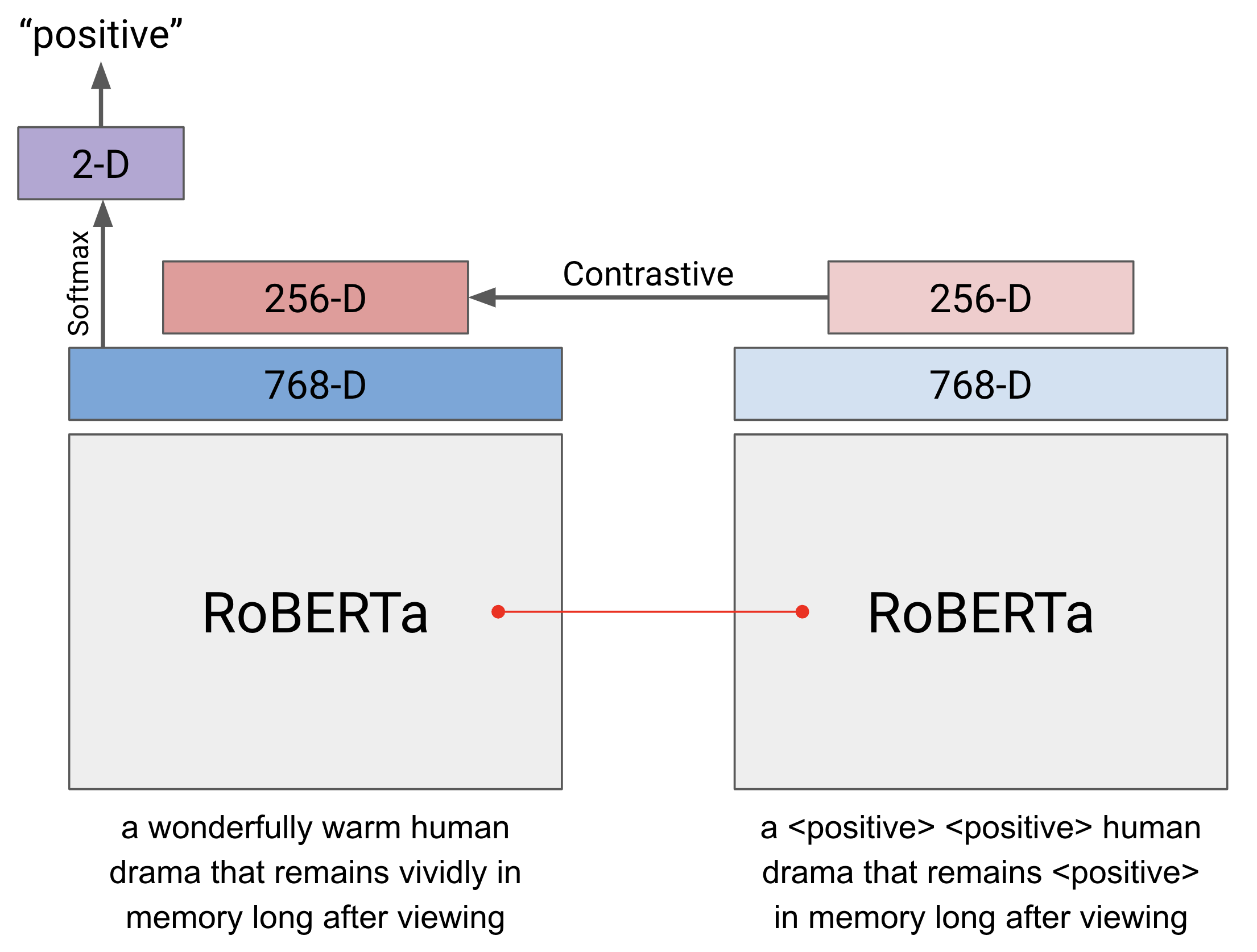}
    \caption{Overview of the proposed method. Although \textsc{RoBERTa} is shown, any pretrained language model will work with this approach. The two RoBERTa networks share the same weights. The dimension of the projection layer is arbitrary.}
    \label{fig:dascl_overview}
\end{figure}

The approach consists of keyword simplification and the contrastive objective function. Figure \ref{fig:dascl_overview} shows an overview of the proposed framework. 

\subsection{Keyword Simplification} 
\label{sec:keyword_simplification}
The first step of the DASCL framework is keyword simplification. We select a set of $M$ dictionaries $\mathcal{D}$. For each dictionary $d_i \in \mathcal{D}$, $i \in \{1,...,M\}$, we assign a token $t_i$. Then, we iterate through the corpus and replace any word $w_j$ in dictionary $d_i$ with the token $t_i$. We repeat these steps for each dictionary. For example, if we have a dictionary of positive words, then applying keyword simplification to
\begin{center}
    \small
    \texttt{a wonderfully warm human drama that remains vividly in memory long after viewing}
\end{center} 
would yield
\begin{center}
    \small 
    \texttt{a <positive> <positive> human drama that remains <positive> in memory long after viewing}
\end{center}

There are many off-the-shelf dictionaries that can be used during keyword simplification. Table \ref{tab:dictionaries_list} in Section \ref{sec:example_dicts} contains a sample of dictionaries reflecting various potential concepts of interest. 

\subsection{Dictionary-Assisted Supervised Contrastive Learning (DASCL) Objective} 
\label{sec:dascl_objective_details}
The dictionary-assisted supervised contrastive learning loss function resembles the loss functions from \citet{khosla_teterwak_wang_tian_isola_maschinot_liu_2020} and  \citet{gunel2020supervisedcontrastive}. Consistent with \citet{khosla_teterwak_wang_tian_isola_maschinot_liu_2020}, we project the final hidden layer of the pretrained language model to an embedding of a lower dimension before using the contrastive loss function.  

Let $\Psi(x_i)$, $i \in \{1,...N\}$, be the $L_2$-normalized projection of the output of the pretrained language encoder for the original text and $\Psi(x_{i+N})$ be the corresponding $L_2$-normalized projection of the output for the keyword-simplified text. $\tau > 0$ is the temperature parameter that controls the separation of the classes, and $\lambda \in [0,1]$ is the parameter that balances the cross-entropy and the DASCL loss functions. We choose $\lambda$ and directly optimize $\tau$ during training. In our experiments, we use the classifier token as the output of the pretrained language encoder. Equation \ref{eq:dascl} is the DASCL loss, Equation \ref{eq:ce} is the multiclass cross-entropy loss, and Equation \ref{eq:overall_loss} is the overall loss that is optimized when fine-tuning the pretrained language model. The original text and the keyword-simplified text are used with the DASCL loss (Eq. \ref{eq:dascl}); only the original text is used with the cross-entropy loss. The keyword-simplified text is not used during inference. 

\begin{small}
\begin{gather}
\mathcal{L}_{\text{DASCL}} = -\frac{1}{2N} \sum_{i=1}^{2N} \frac{1}{2N_{y_i} - 1}  \times \nonumber \\
\sum_{j=1}^{2N} \mathbbm{1}_{i \neq j, y_i=y_j} \log \left[ \frac{\exp(\Psi(x_i) \cdot \Psi(x_j) / \tau)}{\sum_{k=1}^{2N} \mathbbm{1}_{i \neq k} \exp(\Psi(x_i) \cdot \Psi(x_k) / \tau)} \right]
\label{eq:dascl}
\end{gather}
\end{small}

\vspace{-5mm}

\begin{equation}
    \mathcal{L}_{\text{CE}} = -\frac{1}{N} \sum_{i=1}^{N} \sum_{c=0}^{C} y_{i,c} \cdot \log \hat{y}_{i,c}
\label{eq:ce} 
\end{equation}

\vspace{-2mm}

\begin{equation}
    \mathcal{L} = (1-\lambda)\mathcal{L}_{CE} + \lambda\mathcal{L}_{DASCL}
\label{eq:overall_loss}
\end{equation}

\section{Experiments}
\subsection{Few-Shot Learning with SST-2} 
SST-2, a GLUE benchmark dataset \citep{wang-etal-2018-glue}, consists of sentences from movie reviews and binary labels of sentiment (positive or negative). Similar to \citet{gunel2020supervisedcontrastive}, we experiment with SST-2 with three training set sizes: $N{=}$20, 100, and 1,000. Accuracy is this benchmark's primary metric of interest; we also report average precision. We use \textsc{RoBERTa$_{\text{base}}$} as the pretrained language model. For keyword simplification, we use the opinion lexicon \citep{hu_bing_opinionlexicon_2004}, which contains dictionaries of positive and negative words. Section \ref{sec:sst2_dictionaries} further describes these dictionaries. 

We compare DASCL to two other baselines: \textsc{RoBERTa$_{\text{base}}$} using the cross-entropy (CE) loss function and the combination of the cross-entropy and supervised contrastive learning (SCL) loss functions used in \citet{gunel2020supervisedcontrastive}. We also experiment with augmenting the corpus with the keyword-simplified text (referred to as ``data augmentation,'' or ``DA,'' in results tables). In other words, when data augmentation is used, both the original text and the keyword-simplified text are used with the cross-entropy loss. 

We use the original validation set from the GLUE benchmark as the test set, and we sample our own validation set from the training set of equal size to this test set. Further details about the data and hyperparameter configurations can be found in Section \ref{sec:sst2_additional_info}. Table \ref{tab:sst2_fs} shows the results across the three training set configurations. 

\begin{table}
\small
\centering
\addtolength{\tabcolsep}{-2.75pt}  
\begin{tabular}{cccc}
\hline 
\textbf{Loss} & \textbf{$\mathbf{N}$} & \textbf{Accuracy} & \textbf{Avg. Precision}\\ 
\hline 
CE & 20 & $.675 \pm .066$ & $.791 \pm .056$ \\
CE w/ DA & 20 & $.650 \pm .051$ & $.748 \pm .050$ \\
CE+SCL & 20 & $.709 \pm .077$ & $.826 \pm .068$\\ 
CE+DASCL & 20 & $\mathbf{.777 \pm .024}$ & $\mathbf{.871 \pm .014}$\\
CE+DASCL w/ DA & 20 & $.697 \pm .075$ & $.796 \pm .064$\\
\hline
CE & 100 & $.822 \pm .019$ & $.897 \pm .023$  \\
CE w/ DA & 100 & $.831 \pm .032$ & $.904 \pm .031$  \\ 
CE+SCL & 100 & $.833 \pm .042$ & $.883 \pm .043$  \\ 
CE+DASCL & 100 & $\mathbf{.858 \pm .017}$ & $\mathbf{.935 \pm .012}$ \\
CE+DASCL w/ DA & 100 & $.828 \pm .020$ & $.908 \pm .012$  \\
\hline 
CE & 1000 & $.903 \pm .006$ & $\mathbf{.962 \pm .007}$ \\
CE w/ DA & 1000 & $.899 \pm .005$ & $.956 \pm .006$ \\ 
CE+SCL & 1000 & $.905 \pm .005$ & $.960 \pm .011$ \\ 
CE+DASCL & 1000 & $\mathbf{.906 \pm .006}$ & $.959 \pm .009$ \\
CE+DASCL w/ DA & 1000 & $.904 \pm .004$ & $.960 \pm .011$\\ 
\hline 
\end{tabular}
\normalsize
\caption{Accuracy and average precision over the SST-2 test set in few-shot learning settings. Results are averages over 10 random seeds with standard deviations reported. DA refers to \textbf{d}ata \textbf{a}ugmentation, where the keyword-simplified text augments the training corpus.} 
\label{tab:sst2_fs} 
\end{table} 

DASCL improves results the most when there are only a few observations in the training set. When $N{=}20$, using DASCL yields a $10.2$ point improvement in accuracy over using the cross-entropy loss function ($p{<}.001$) and a $6.8$ point improvement in accuracy over using the SCL loss function ($p{=}.023$). Figure \ref{fig:n20_sst2_tsne} in Section \ref{sec:tsne_plots_N20} visualizes the learned embeddings using each of these loss functions using t-SNE plots. When the training set's size increases, the benefits of using DASCL decrease. DASCL only has a slightly higher accuracy when using 1,000 labeled observations, and the difference between DASCL and cross-entropy alone is insignificant ($p{=}.354$). 

\subsection{\textit{New York Times} Articles about the Economy} 
\citet{barbera_boydstun_linn_mcmahon_nagler_2021} classify the tone of \textit{New York Times} articles about the American economy as positive or negative. 3,119 of the 8,462 labeled articles (3,852 unique articles) in the training set are labeled positive; 162 of the 420 articles in the test set are labeled positive. Accuracy is the primary metric of interest; we also report average precision. In addition to using the full training set, we also experiment with training sets of sizes 100 and 1,000. We use the positive and negative dictionaries from Lexicoder \citep{young_soroka_2012} and dictionaries of positive and negative economic terms \citep{hopkins_kim_kim_2017_newspapercoverage}. \citet{barbera_boydstun_linn_mcmahon_nagler_2021} use logistic regression with $L_2$ regularization. We use \textsc{RoBERTa$_{\text{base}}$} as the pretrained language model. Section \ref{sec:econmedia_additional_info} contains more details about the data, hyperparameters, and other evaluation metrics. Table \ref{tab:econmedia_results} shows the results across the three training set configurations. 

\begin{table}
    \small 
    \centering
    \addtolength{\tabcolsep}{-2.75pt}
    \begin{tabular}{cccc}
    \hline 
    \textbf{Loss} & $\mathbf{N}$ & \textbf{Accuracy} & \textbf{Avg. Precision} \\ 
    \hline
    L2 Logit & 100 & $.614$ & $.479$ \\ 
    CE & 100 & $.673 \pm .027$ & $.593 \pm .048$   \\
    CE w/ DA & 100 & $.663 \pm .030$ & $.576 \pm .058$ \\
    CE+SCL & 100 & $.614 \pm .000$ & $.394 \pm .043$ \\ 
    CE+DASCL & 100 & $.705 \pm .013$ & $\mathbf{.645 \pm .016}$ \\
    CE+DASCL w/ DA & 100 & $\mathbf{.711 \pm .013}$ & $.644 \pm .027$ \\ 
    \hline 
    L2 Logit & 1000 & $.624$ & $.482$ \\ 
    CE & 1000 & $.716 \pm .012$ & $.662 \pm .030$ \\ 
    CE w/ DA & 1000 & $.710 \pm .011$ & $.656 \pm .024$  \\ 
    CE+SCL & 1000 & $.722 \pm .009$ & $.670 \pm .022$ \\
    CE+DASCL & 1000 & $.732 \pm .011$ & $.671 \pm .025$ \\
    CE+DASCL w/ DA & 1000 & $\mathbf{.733 \pm .008}$ & $\mathbf{.681 \pm .021}$ \\
    \hline 
    L2 Logit & Full & $.681$ & $.624$ \\
    CE & Full &  $.753 \pm .012$ & $.713 \pm .015$ \\
    CE w/ DA & Full & $.752 \pm .011$ & $.708 \pm .017$ \\
    CE+SCL & Full & $.756 \pm .011$ & $.723 \pm .009$ \\ 
    CE+DASCL & Full & $.759 \pm .006$ & $\mathbf{.741 \pm .010}$ \\
    CE+DASCL w/ DA & Full & $\mathbf{.760 \pm .008}$ & $.739 \pm .014$  \\
    \hline 
    \end{tabular}
    \caption{Accuracy and average precision over the economic media test set \citep{barbera_boydstun_linn_mcmahon_nagler_2021} when using 100, 1000, and all labeled examples from the training set for fine-tuning. Except for logistic regression, results are averages over 10 random seeds with standard deviations reported.}  
    \label{tab:econmedia_results}
\end{table}

When $N{=}100$, DASCL outperforms cross-entropy only, cross-entropy with data augmentation, and SCL on accuracy ($p{<}.005$ for all) and average precision ($p{<}.01$ for all). When $N{=}1000$, DASCL outperforms cross-entropy only, cross-entropy with data augmentation, and SCL on accuracy ($p{<}.05$ for all) and average precision (but not statistically significantly). DASCL performs statistically equivalent to DASCL with data augmentation across all metrics when $N{=}100$ and $1000$. 

When using the full training set, \textsc{RoBERTa$_{\text{base}}$} is a general improvement over logistic regression. Although the DASCL losses have slightly higher accuracy than the other RoBERTa-based models, the differences are not statistically significant. Using DASCL yields a $2.8$ point improvement in average precision over cross-entropy ($p{<}.001$) and a $1.8$ improvement in average precision over SCL ($p{<}.001$). Figure \ref{fig:econmedia_tsne} in Section \ref{sec:tsne_plots_econmedia} visualizes the learned embeddings using each of these loss functions using t-SNE plots.  

\subsection{Abusive Speech on Social Media}
The OffensEval dataset \citep{zampieri-etal-2020-semeval} contains 14,100 tweets annotated for offensive language. A tweet is considered offensive if ``it contains any form of non-acceptable language (profanity) or a targeted offense.'' \citet{caselli-etal-2020-feel} used this same dataset and more narrowly identified tweets containing ``hurtful language that a speaker uses to insult or offend another individual or group of individuals based on their personal qualities, appearance, social status, opinions, statements, or actions.'' We focus on this dataset, AbusEval, because of its greater conceptual difficulty. 2,749 of the 13,240 tweets in the training set are labeled abusive, and 178 of the 860 tweets in the test set are labeled abusive. \citet{caselli-etal-2021-hatebert} pretrain a \textsc{BERT$_{\text{base-uncased}}$} model, HateBERT, using the Reddit Abusive Language English dataset. Macro F1 and F1 over the positive class are the primary metrics of interest; we also report average precision. In addition to using the full training set, we also experiment with training sets of sizes 100 and 1,000. Section \ref{sec:abuseval_additional_info} contains more details about the data and hyperparameters.

We combine DASCL with \textsc{BERT$_{\text{base-uncased}}$} and HateBERT. \citet{alorainy_othering} detect cyber-hate speech using threats-based othering language, focusing on the use of ``us'' and ``them'' pronouns. Following their strategy, we look at the conjunction of sentiment using Lexicoder and two dictionaries of ``us'' and ``them'' pronouns, which may suggest abusive speech. Table \ref{tab:abuseval_results} compares the performance of \textsc{BERT$_{\text{base-uncased}}$} and HateBERT with cross-entropy against \textsc{BERT$_{\text{base-uncased}}$} and HateBERT with cross-entropy and DASCL. 

\begin{table}
    \small
    \centering
    \addtolength{\tabcolsep}{-4.5pt}  
    \begin{tabular}{ccccc}
    \hline 
    \textbf{Model} & $\mathbf{N}$ & \textbf{Macro F1} & \textbf{F1, Pos} & \textbf{Avg. Precision}\\
    \hline
    \footnotesize BERT & \footnotesize 100 & $.293 {\pm} .083$ & $.334 {\pm} .025$ & $.233 {\pm} .040$ \\ 
    \footnotesize HateBERT & \footnotesize 100 & $\mathbf{.513 {\pm} .120}$ & $.346 {\pm} .049$ & $\mathbf{.303 {\pm} .059}$  \\
    \begin{tabular}[c]{@{}c@{}}\footnotesize BERT \\ \footnotesize w/ DASCL\end{tabular} & \footnotesize 100 & $.427 {\pm} .112$ & $\mathbf{.362 {\pm} .017}$ & $.284 {\pm} .022$ \\
    \begin{tabular}[c]{@{}c@{}}\footnotesize HateBERT \\ \footnotesize w/ DASCL\end{tabular} & \footnotesize 100 & $.449 {\pm} .110$ & $.345 {\pm} .033$ & $.281 {\pm} .045$ \\
    \hline 
    \footnotesize BERT & \footnotesize 1000 & $.710 {\pm} .018$ & $.523 {\pm} .034$ & $.608 {\pm} .024$ \\ 
    \footnotesize HateBERT & \footnotesize 1000 & $.711 {\pm} .028$ & $.512 {\pm} .053$ & $.626 {\pm} .017$  \\
    \begin{tabular}[c]{@{}c@{}}\footnotesize BERT \\ \footnotesize w/ DASCL\end{tabular} & \footnotesize 1000 & $\mathbf{.729 {\pm} .014}$ & $\mathbf{.559 {\pm} .032}$ & $\mathbf{.637 {\pm} .016}$ \\
    \begin{tabular}[c]{@{}c@{}}\footnotesize HateBERT \\ \footnotesize w/ DASCL\end{tabular} & \footnotesize 1000 & $.704 {\pm} .010$ & $.501 {\pm} .020$ & $.626 {\pm} .016$ \\
    \hline
    \footnotesize BERT & \footnotesize Full & $.767 {\pm} .008$ & $.636 {\pm} .012$ & $.703 {\pm} .006$ \\ 
    \footnotesize HateBERT & \footnotesize Full &  $.768 {\pm} .005$ & $.630 {\pm} .009$ & $.695 {\pm} .005$ \\
    \begin{tabular}[c]{@{}c@{}}\footnotesize BERT \\ \footnotesize w/ DASCL\end{tabular} & \footnotesize Full &  $\mathbf{.779 {\pm} .010}$ & $\mathbf{.653 {\pm} .014}$ & $\mathbf{.716 {\pm} .005}$ \\
    \begin{tabular}[c]{@{}c@{}}\footnotesize HateBERT \\ \footnotesize w/ DASCL\end{tabular} & Full & $.766 {\pm} .007$ & $.623 {\pm} .011$ & $.695 {\pm} .006$ \\
    \hline 
    \end{tabular}
    \caption{Macro F1, F1, and average precision over the AbusEval test set \citep{caselli-etal-2020-feel} when using 100, 1000, and all labeled examples from the training set for fine-tuning. Results are averages over 10 random seeds with standard deviations reported.}
    \label{tab:abuseval_results}
\end{table}

When $N{=}100$, BERT with DASCL outperforms BERT on macro F1 ($p{=}.008$), F1 over the positive class ($p{=}.011$), and average precision ($p{=}.003$); when $N{=}1000$, BERT with DASCL outperforms BERT on macro F1 ($p{=}.021$), F1 over the positive class ($p{=}.028$), and average precision ($p{=}.007$). HateBERT with DASCL performs statistically on par with HateBERT across all metrics for $N{=}100$ and $N{=}1000$. BERT with DASCL performs statistically equivalent to HateBERT when $N{=}100$ and $N{=}1000$ on all metrics, except on F1 over the positive class when $N{=}1000$ ($p{=}.030$).  

When using the full training set, BERT with DASCL improves upon the macro F1, F1 over the positive class, and average precision compared with both BERT (macro F1: $p{=}.010$; F1: $p{=}.010$; AP: $p{<}.001$) and HateBERT (macro F1: $p{=}.007$; F1: $p{<}.001$; AP: $p{<}.001$). Figure \ref{fig:abuseval_tsne} in Section \ref{sec:tsne_plots_abuseval} visualizes the learned embeddings using BERT and BERT with DASCL using t-SNE plots.  

\section{Conclusion} 
We propose a supervised contrastive learning approach that allows researchers to leverage specialized dictionaries when fine-tuning pretrained language models. We show that using DASCL with cross-entropy improves classification performance on SST-2 in few-shot learning settings, on classifying perceptions about the economy expressed in \textit{New York Times} articles, and on identifying tweets containing abusive language when compared to using cross-entropy alone or alternative contrastive and data augmentation methods. In the future, we aim to extend our approach to other supervised contrastive learning frameworks, such as using this method to upweight difficult texts \citep{suresh-ong-2021-negatives}. We also plan to expand this approach to semi-supervised and self-supervised settings to better understand core concepts expressed in text. 

\section*{Limitations}
We aim to address limitations to the supervised contrastive learning approach described in this paper in future works. We first note that there are no experiments in this paper involving multiclass or multilabel classification; all experiments involve only binary classification. Multiclass or multilabel classification may present further challenges when categories are more nuanced. We expect improvements in classification performance when applied to multiclass or multilabel classification settings, but we have not confirmed this. 

Second, we have not experimented with the dimensionality of the projection layer or the batch sizes. At the moment, the projection layer is arbitrarily set to 256 dimensions, and we use batch sizes from previous works. Future work aims to study how changing the dimensionality of this projection layer and the batch size affects classification outcomes. 

Third, we have used the DASCL objective with \textsc{RoBERTa} and \textsc{BERT}, but have not used it with the latest state-of-the-art pretrained language models. We focused on these particular pretrained language models because they are commonly used in the social sciences and because of computational constraints.

Fourth, we have not examined how the quality or size of the dictionary may affect classification outcomes. A poorly constructed dictionary may lead to less improvement in classification performance metrics or may even hurt performance. Dictionaries with too many words or too few words may also not lead to improvements in classification performance metrics. Future work aims to study how the quality and size of dictionaries affect the DASCL approach. 

Fifth, we have not explored how this method can be used to potentially reduce bias in text classification. For example, we can replace gendered pronouns with a token (such as ``$<$pronoun$>$''), potentially reducing gender bias in analytical contexts such as occupation. 

Lastly, we have not explored how keyword simplification may be useful in a self-supervised or semi-supervised contrastive learning setting. This may be particularly helpful for social scientists who are often interested in exploring core concepts or perspectives in text rather than classifying text into specific classes. 

\section*{Ethics Statement}
Our paper describes a supervised contrastive learning approach that allows researchers to leverage specialized dictionaries when fine-tuning pretrained language models. While we did not identify any systematic biases in the particular set of dictionaries we used, any dictionary may encode certain biases and/or exclude certain groups. This can be particularly problematic when working with issues such as detecting hate speech and abusive language. For example, in the context of abusive language, if words that attack a particular group are (purposely or unintentionally) excluded from the dictionaries, those words would not be replaced. This may under-detect abusive text that attacks this specific group.  

This paper does not create any new datasets. The OffensEval/AbusEval dataset contains sensitive and harmful language. Although we did not annotate or re-annotate any tweets, we are cognizant that particular types of abusive language against certain groups or identities may not have been properly annotated as abusive, or certain types of abusive language may have been excluded from the corpus entirely. 

\section*{Acknowledgements}
We gratefully acknowledge that the Center for Social Media and Politics at New York University is supported by funding from the John S. and James L. Knight Foundation, the Charles Koch Foundation, Craig Newmark Philanthropies, the William and Flora Hewlett Foundation, the Siegel Family Endowment, and the Bill and Melinda Gates Foundation. This work was supported in part through the NYU IT High Performance Computing resources, services, and staff expertise. We thank the members of the Center for Social Media and Politics for their helpful comments when workshopping this paper. We would also like to thank the anonymous reviewers for their valuable feedback in improving this paper.

\bibliography{anthology,custom}
\bibliographystyle{acl_natbib}

\appendix
\section{Appendix}
\label{sec:appendix}


\subsection{\citet{gunel2020supervisedcontrastive}'s Supervised Contrastive Learning Objective}
\label{sec:gunelscllossfunction}
Equation \ref{eq:scl} is the supervised contrastive learning objective from \citet{gunel2020supervisedcontrastive}. The dictionary-assisted supervised contrastive learning objective in Equation \ref{eq:dascl} is similar to Equation \ref{eq:scl} except Equation \ref{eq:dascl} is extended to include the keyword-simplified text. 

\begin{small}
\begin{gather}
\mathcal{L}_{SCL} = \sum_{i=1}^{N} -\frac{1}{N_{y_i} - 1} \times \nonumber \\ \sum_{j=1}^{N} \mathbbm{1}_{i \neq j} \mathbbm{1}_{y_i = y_j} \log \left[ \frac{\exp(\Phi(x_i) \cdot \Phi(x_j) / \tau)}{\sum_{k=1}^{N} \mathbbm{1}_{i \neq k} \exp(\Phi(x_i) \cdot \Phi(x_k) / \tau)} \right]
\label{eq:scl}
\end{gather}
\end{small}

\subsection{Examples of Dictionaries and Lexicons}
\label{sec:example_dicts} 
Table \ref{tab:dictionaries_list} contains a sample of dictionaries across various use cases and academic fields that can be potentially used with DASCL. There is no particular order to the dictionaries, with similar dictionaries clustered together. We did not include any non-open source dictionaries. 

\begin{table*}
\centering
\small
\addtolength{\tabcolsep}{-3pt}
\begin{tabular}{c|c|c}
\textbf{Dictionary Name} & \textbf{Type of Words or Phrases} & \textbf{Source}\\ 
\hline 
Opinion Lexicon & Positive/negative sentiment & \citet{hu_bing_opinionlexicon_2004} \\
Lexicoder Sentiment Dictionary & Positive/negative sentiment & \citet{young_soroka_2012} \\
SentiWordNet & Sentiment & \citet{baccianella-etal-2010-sentiwordnet} \\ 
ANEW & Emotions & \citet{ANEW} \\ 
EmoLex & Emotions & \citet{Mohammad13} \\
DepecheMood & Emotions & \citet{staiano-guerini-2014-depeche} \\
Moodbook & Emotions & \citet{moodbook} \\ 
Moral \& Emotion Dictionaries & Moralization and Emotions & \citet{brady_wills_jost_tucker_bavel_2017} \\
Emotion Intensity Lexicon & Emotion intensity & \citet{mohammad-2018-word} \\
VAD Lexicon & \scriptsize{Valence, arousal, and dominance of words} & \citet{mohammad-2018-obtaining} \\
SCL-NMA & Negators, modals, degree adverbs & \citet{kiritchenko-mohammad-2016-capturing} \\
SCL-OPP & Sentiment of mixed polarity phrases & \citet{kiritchenko-mohammad-2016-sentiment} \\ 
Dictionary of Affect in Language & Affect of English words & \citet{WHISSELL1989113,whissell_updated} \\
Spanish DAL & Affect of Spanish words & \scriptsize{\citet{dell-amerlina-rios-gravano-2013-spanish}} \\
Subjectivity Lexicon & Subjectivity clues & \citet{wilson-etal-2005-recognizing} \\
Subjectivity Sense Annotations & Subjectivity disambiguation & \citet{wiebe-mihalcea-2006-word} \\
Arguing Lexicon & Patterns representing arguing & \citet{somasundaran-etal-2007-detecting} \\
+/-Effect Lexicon & Positive/negative effects on entities & \citet{choi-wiebe-2014-effectwordnet} \\
AEELex & Arabic and English emotion & \citet{shakil_etal_aeelex} \\ 
Discrete Emotions Dictionary & Emotions in news content & \citet{Fioroni_Hasell_Soroka_Weeks_2022} \\ 
Yelp/Amazon Reviews Dictionaries & Sentiments in reviews & \citet{kiritchenko-etal-2014-nrc} \\ 
VADER & Sentiments on social media & \citet{Hutto_Gilbert_2014} \\
English Twitter Sentiment Lexicon & Sentiments on social media & \citet{rosenthal-etal-2015-semeval} \\
Arabic Twitter Sentiment Lexicon & Sentiments on social media & \citet{kiritchenko-etal-2016-semeval} \\ 
Hashtag Emotion Lexicon & Emotions associated with Twitter hashtags & \citet{hashtags_emotions} \\  
Political Polarization Dictionary & Political polarizing language & \citet{simchon_brady_vanbavel_2022_trollanddivide} \\
ed8 & Affective language in German political text & \citet{widmann_wich_2022} \\ 
Policy Agendas Dictionary & Keywords by policy area & \citet{policy_agendas} \\ 
``Women'' Terms & Terms related to the category ``women'' & \citet{pearson_dancey_2011} \\ 
Trump's Twitter Insults & Insults used by President Trump & \citet{trump_insults} \\
Hate Speech Lexicon & Hate speech & \citet{Davidson_Warmsley_Macy_Weber_2017} \\
\scriptsize{PeaceTech Lab's Hate Speech Lexicons} & Hate speech & \citet{peacetech_hate_speech_lexicon} \\
Hatebase & Hate speech & \citet{hatebase} \\ 
Hurtlex & Hate speech & \citet{hurtlex} \\ 
Hate on Display Hate Symbols & Hate speech and symbols & \citet{adl_symbols} \\ 
Hate speech on Twitter & Hate speech on Twitter & \citet{QJPS-19045} \\ 
Reddit hate lexicon & Hate speech on Reddit & \citet{reddit_hate} \\ 
Pro/Anti-Lynching & Pro/anti-lynching terms & \citet{weaver_2019} \\
Grievance Dictionary & Terms related to grievance & \citet{vandervegt2020grievance} \\ 
Economic Sentiment Terms & Economic sentiment in newspapers & \citet{hopkins_kim_kim_2017_newspapercoverage} \\ 
Loughran-McDonald Dictionary & Financial sentiment & \citet{loughran_mcdonald_financial} \\
SentiEcon & Financial and economic sentiment & \citet{moreno-ortiz-etal-2020-design} \\
Financial Phrase Bank & Phrases expressing financial sentiment & \citet{Malo2014GoodDO} \\
Stock Market Sentiment Lexicon & Stock market sentiment & \citet{OLIVEIRA201662} \\
BioLexicon & \scriptsize{Linguistic information of biomedical terms} & \citet{Thompson2011} \\
SentiHealth & Health-related sentiment & \citet{Asghar2016} \\ 
MEDLINE Abbreviations & Abbreviations from medical abstracts & \citet{medline_abbreviations} \\ 
COVID-CORE Keywords & COVID-19 keywords on Twitter & \citet{covidcoreumich} \\
PSi Lexicon & Performance studies keywords & \citet{psi_lexicon} \\ 
Concreteness Ratings & Concreteness of words and phrases & \citet{Brysbaert2013} \\ 
Word-Colour Association Lexicon & Word-color associations & \citet{mohammad-2011-even} \\
Regressive Imagery Dictionary & Primordial vs. conceptual thinking & \citet{martindale1975romantic} \\ 
Empath & Various lexical categories & \citet{empath_lexicon} \\
WordNet & Various lexical categories & \citet{wordnet} 
\end{tabular}
\normalsize
\caption{A sample of dictionaries that can potentially be used with DASCL. There is no particular order to the dictionaries, with similar dictionaries clustered together. We did not include any non-open source dictionaries.} 
\label{tab:dictionaries_list} 

\end{table*}

\subsection{Additional Information for the Few-Shot Learning Experiments with SST-2}
\label{sec:sst2_additional_info} 

\subsubsection{Data Description: Few-Shot Training Sets, Validation Set, and Test Set} 
The SST-2 dataset was downloaded using Hugging Face's Datasets library \citep{lhoest-etal-2021-datasets}. The test set from SST-2 does not contain any labels, so we use the validation set from SST-2 as our test set. We create our own validation set by randomly sampling a dataset equivalent in size to the original validation set. Our validation set and few-shot learning sets were sampled with no consideration to the label distributions of the original training or validation sets. 

When $N=20$, there are 11 positive examples and 9 negative examples. When $N=100$, there are 60 positive examples and 40 negative examples. When $N=1000$, there are 558 positive examples and 442 negative examples. Our validation set has 486 positive examples and 386 negative examples. Lastly, our test set has 444 positive examples and 428 negative examples.  

\subsubsection{Text Preprocessing Steps}
\label{sec:sst2_preprocessing} 
The only text preprocessing step taken is that non-ASCII characters are removed from the dataset. The text is tokenized using a byte-level BPE tokenizer \citep{liu2019roberta}. 

\subsubsection{Dictionaries Used During Keyword Simplification} 
\label{sec:sst2_dictionaries} 
We used the opinion lexicon from \citet{hu_bing_opinionlexicon_2004}. This lexicon consists of two dictionaries: one with all positive unigrams and one with all negative unigrams. There are 2,006 positive words and 4,783 negative words. We replaced the positive words with the token ``$<$positive$>$''. We replaced the negative words with the token ``$<$negative$>$''. 

\subsubsection{Number of Parameters and Runtime} 
\label{sec:sst2_params_runtime} 
This experiment uses the \textsc{RoBERTa$_{\text{base}}$} pretrained language model, which contains 125 million parameters \citep{liu2019roberta}. When using DASCL, we also had an additional temperature parameter, $\tau$, that was directly optimized. With the hyperparameters described in Section \ref{sec:sst2_hyperparam_configs} and using an NVIDIA V100 GPU, it took approximately 2.1 seconds to train over 40 batches using cross-entropy (CE) alone, 2.2 seconds to train over 40 batches using CE+SCL, and 3.3 seconds to train over 40 batches using CE+DASCL. 

\subsubsection{Hyperparameter Selection and Configuration Details}
\label{sec:sst2_hyperparam_configs}
We take our hyperparameter configuration directly from \citet{gunel2020supervisedcontrastive}. For each configuration, we set the learning rate to $1 \times 10^{-5}$ and used a batch size of $16$. When using the SCL objective, in line with \citet{gunel2020supervisedcontrastive}, we set $\lambda = 0.9$ and $\tau = 0.3$. When using the DASCL objective, we also set $\lambda = 0.9$ and initialized $\tau = 0.3$. We trained for 100 epochs for all few-shot learning settings. 

\subsubsection{Model Evaluation Details} 
\label{sec:sst2_model_eval_details}
The model from the epoch with the highest accuracy over our own validation set was chosen as the final model for each random seed. We report accuracy, which is the main metric of interest with this benchmark, and average precision. Average precision is used to summarily quantify the precision-recall tradeoff, and is viewed as the area under the precision-recall curve \citep{davis_2006_ap}. Average precision is defined as 
\[AP = \sum_{n} \left(R_n - R_{n-1}\right) P_n\]
where $P_n$ and $R_n$ are the precision and recall at the $n$th threshold. 

\subsubsection{Results over the Validation Set} 
\label{sec:sst2_validation_results}
Table \ref{tab:sst2_fs_validation} reports the accuracy and the average precision over the validation set for the SST-2 few-shot setting experiments. The validation set was used for model selection, so the reported results over the validation set are from the model with the highest accuracy achieved on the validation set across the $100$ epochs.

\begin{table}
\small
\centering
\addtolength{\tabcolsep}{-2.5pt}  
\begin{tabular}{cccc}
\hline 
\textbf{Loss} & \textbf{$\mathbf{N}$} & \textbf{Accuracy} & \textbf{Avg. Precision}\\ 
\hline 
CE & 20 & $.680 \pm .043$ & $.768 \pm .062$ \\
CE w/ DA & 20 & $.668 \pm .025$ & $.729 \pm .034$ \\
CE+SCL & 20 & $.707 \pm .049$ & $.797 \pm .060$  \\ 
CE+DASCL & 20 & $.743 \pm .016$ & $.839 \pm .023$\\
CE+DASCL w/ DA & 20 & $.700 \pm .048$ & $.765 \pm .061$ \\
\hline
CE & 100 & $.832 \pm .015$ & $.905 \pm .021$ \\
CE w/ DA & 100 & $.849 \pm .023$ & $.915 \pm .027$ \\ 
CE+SCL & 100 & $.848 \pm .020$ & $.905 \pm .029$  \\ 
CE+DASCL & 100 & $.872 \pm .010$ & $.942 \pm .010$\\
CE+DASCL w/ DA & 100 & $.842 \pm .020$ & $.927 \pm .010$ \\
\hline 
CE & 1000 & $.900 \pm .005$ & $.958 \pm .010$ \\
CE w/ DA & 1000 & $.905 \pm .006$ & $.959 \pm .005$ \\ 
CE+SCL & 1000 & $.904 \pm .038$ & $.958 \pm .016$ \\ 
CE+DASCL & 1000 & $.907 \pm .004$ & $.961 \pm .011$ \\
CE+DASCL w/ DA & 1000 & $.908 \pm .005$ & $.965 \pm .008$ \\ 
\hline 
\end{tabular}
\normalsize
\caption{Accuracy and average precision over the SST-2 validation set in few-shot learning settings. Results are averages over 10 random seeds with standard deviations reported.} 
\label{tab:sst2_fs_validation} 
\end{table} 

\subsubsection{t-SNE Plots of the Learned Classifier Token Embeddings for the Test Set, $\mathbf{N=20}$}
\label{sec:tsne_plots_N20}
\begin{figure*}
\centering
\begin{subfigure}{.45\textwidth}
  \centering
  \includegraphics[width=\linewidth]{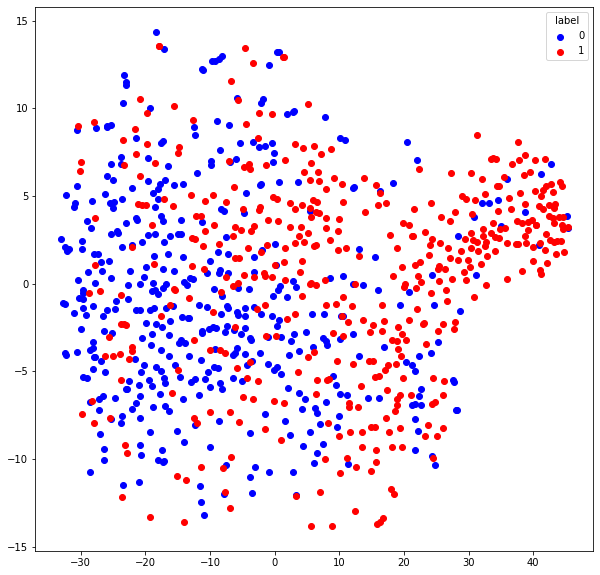}
  \caption{CE}
\end{subfigure}
\begin{subfigure}{.45\textwidth}
  \centering
  \includegraphics[width=\linewidth]{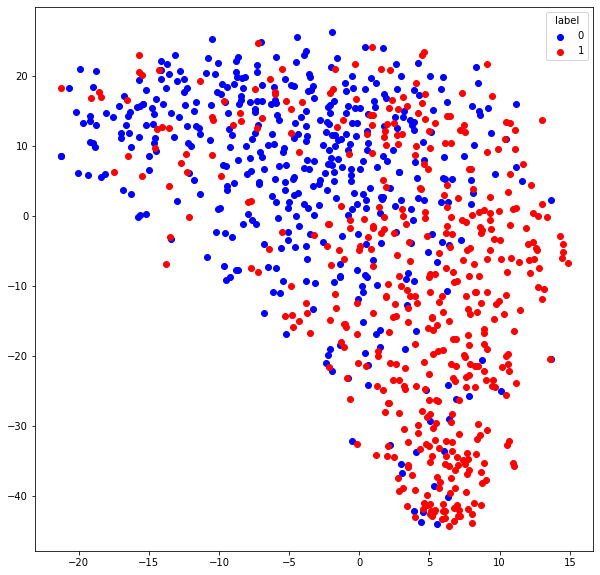}
  \caption{CE+SCL}
\end{subfigure}
\begin{subfigure}{.45\textwidth}
  \centering
  \includegraphics[width=\linewidth]{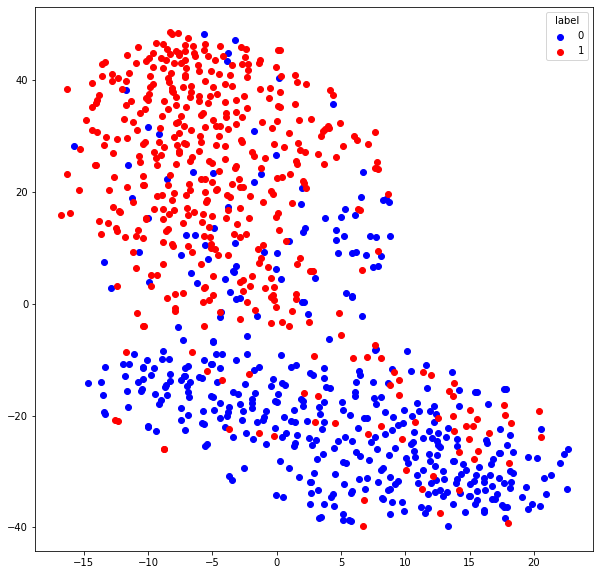}
  \caption{CE+DASCL}
\end{subfigure}
\caption{t-SNE plots of the classifier token embeddings on the SST-2 test set fine-tuned using a training set size of 20. The loss configuration is noted below each plot. Blue are negative examples and red are positive examples.}
\label{fig:n20_sst2_tsne}
\end{figure*}

We use t-SNE \citep{JMLR:v9:vandermaaten08a} plots to visualize the learned classifier token embeddings, ``$<$s$>$'', over the SST-2 test set when using the cross-entropy objective alone, using the cross-entropy objective with the supervised contrastive learning (SCL) objective \citep{gunel2020supervisedcontrastive}, and using the cross-entropy objective with the dictionary-assisted supervised contrastive learning (DASCL) objective. These plots are in Figure \ref{fig:n20_sst2_tsne}. We see that DASCL draws embeddings of the same class closer and pushes embeddings of different classes farther apart compared to using cross-entropy alone or using cross-entropy with SCL.  

\subsection{Additional Information for the \textit{New York Times} Articles about the Economy Experiments} 
\label{sec:econmedia_additional_info}

\subsubsection{Data Description: Few-Shot Training Set, Validation Set, and Test Set} 
The data for the \textit{New York Times} articles was downloaded from \url{https://dataverse.harvard.edu/dataset.xhtml?persistentId=doi:10.7910/DVN/MXKRDE}. The test set was created using the replication files included at the link. In the original code, there was an error with overlapping training and test sets. We removed the duplicated observations from the training set. Because a single article could be annotated multiple times by different annotators, our validation set was created using 15\% of the unique number of articles in the training data. 452 of the 1,317 labeled articles in the validation set are labeled positive. For our few-shot training sets, when $N=100$, there were 41 positive examples. When $N=1000$, there were 363 positive examples. Our validation set and few-shot learning sets were sampled with no consideration to the label distributions of the original training or validation sets. 

\subsubsection{Text Preprocessing Steps} 
The only preprocessing was removing HTML tags that occasionally appeared in the text. The text is tokenized using a byte-level BPE tokenizer. 

\subsubsection{Dictionaries Used During Keyword Simplification} 
\label{sec:econmedia_dictionaries}
We used two sets of dictionaries during keyword simplification. We first used Lexicoder, downloaded from \url{http://www.snsoroka.com/data-lexicoder/}. It is a dictionary specifically designed to study sentiment in news coverage \citep{young_soroka_2012}. The dictionary is split into four separate sub-dictionaries: positive words, negative words, ``negative'' positive words (e.g., ``not great''), and ``negative'' negative words (e.g., ``not bad''). There are 1,709 positive words, 2,858 negative words, 1,721 negative positive words, and 2,860 negative negative words. We replaced positive words and negative negative words with the token ``$<$positive$>$''. We replaced negative words and negative positive words with the token ``$<$negative$>$''. 

The second dictionary we used was the 21 economic terms from \citet{hopkins_kim_kim_2017_newspapercoverage}. The 6 positive economic terms (in stemmed form) are ``bull*'', ``grow*'', ``growth*'', ``inflat*'', ``invest*'', and ``profit*''. The 15 negative economic terms (in stemmed form) are ``bad*'', ``bear*'', ``debt*'', ``drop*'', ``fall*'', ``fear*'', ``jobless*'', ``layoff*'', ``loss*'', ``plung*'', ``problem*'', ``recess*'', ``slow*'', ``slump*'', and ``unemploy*''. We replaced the positive economic words with the token ``$<$positive\_econ$>$''. We replaced the negative economic words with the token ``$<$negative\_econ$>$''.  

\subsubsection{Number of Parameters and Runtime} 
\label{sec:econmedia_numparams_runtime}
This experiment uses the \textsc{RoBERTa$_{\text{base}}$} pretrained language model, which contains 125 million parameters \citep{liu2019roberta}. When using DASCL, we also had an additional temperature parameter, $\tau$, that was directly optimized. With the hyperparameters described in Section \ref{sec:econmedia_hyperparam_configs} and using an NVIDIA V100 GPU, it took approximately 5.7 seconds to train over 40 batches using cross-entropy (CE) alone, 5.7 seconds to train over 40 batches using CE+SCL, and 10.7 seconds to train over 40 batches using CE+DASCL. 

\subsubsection{Hyperparameter Selection and Configuration Details}
\label{sec:econmedia_hyperparam_configs} 
We selected hyperparameters using the validation set. We searched over the learning rate and the temperature initialization; we used $\lambda = 0.9$ for all loss configurations involving contrastive learning. We used a batch size of $8$ because of resource constraints. We fine-tuned \textsc{RoBERTa$_{\text{base}}$} for 5 epochs. 

For the learning rate, we searched over $\{5 \times 10^{-6}, 1 \times 10^{-5}, 2 \times 10^{-5}\}$; for the temperature, $\tau$, initialization, we searched over $\{0.07, 0.3\}$. We fine-tuned the model and selected the model from the epoch with the highest accuracy. We repeated this with three random seeds, and selected the hyperparameter configuration with the highest average accuracy. We used accuracy as the criterion because \citet{barbera_boydstun_linn_mcmahon_nagler_2021} used accuracy as the primary metric of interest. The final learning rate across all loss configurations was $5 \times 10^{-6}$. The final $\tau$ initialization for both SCL and DASCL loss configurations was $0.07$. We used these same hyperparameters when we limited the training set to 100 and 1,000 labeled examples. 

\subsubsection{Model Evaluation Details} 
\label{sec:econmedia_model_eval_details}
During fine-tuning, the model from the epoch with the highest accuracy over the validation set was chosen as the final model for each random seed. We report accuracy, which is the main metric of interest with this dataset, and average precision. For a definition of average precision, see Section \ref{sec:sst2_model_eval_details}. 

The results in Table \ref{tab:econmedia_results} for logistic regression using the full training set differ slightly from their paper because of an error in overlapping training and test sets in the original splits.

\subsubsection{Additional Classification Metrics: Precision and Recall} 
\label{sec:econmedia_precisionrecall}

Table \ref{tab:econmedia_results_test_precision_recall} contains additional classification metrics---precision and recall---for the test set when using 100, 1,000, and all labeled examples from the training set for fine-tuning. 

\begin{table}
    \centering
    \small
    \begin{tabular}{cccc}
    \hline 
    \textbf{Loss} & $\mathbf{N}$ & \textbf{Precision} & \textbf{Recall} \\ 
    \hline 
    L2 Logit & $100$ & $.000$ & $.000$ \\ 
    CE & $100$ & $.646 \pm .079$ & $.359 \pm .126$  \\
    CE+DA & $100$ & $.656 \pm .096$ & $.312 \pm .170$  \\
    SCL & $100$ & $.000 \pm .000$ & $.000 \pm .000$  \\ 
    DASCL & $100$ & $.653 \pm .044$ & $.519 \pm .055$ \\
    DASCL+DA & $100$ & $.690 \pm .051$ & $.475 \pm .093$\\
    \hline
    L2 Logit & $1000$ & $.542$ & $.160$ \\ 
    CE & $1000$ & $.670 \pm .028$ & $.526 \pm .074$  \\
    CE+DA & $1000$ & $.658 \pm .029$ & $.527 \pm .089$  \\
    SCL & $1000$ & $.674 \pm .017$ & $.543 \pm .046$  \\ 
    DASCL & $1000$ & $.684 \pm .032$ & $.575 \pm .069$ \\
    DASCL+DA & $1000$ & $.682 \pm .036$ & $.592 \pm .062$\\
    \hline 
    L2 Logit & Full & $.689$ & $.315$\\
    CE & Full & $.690 \pm .036$  & $.663 \pm .055$ \\
    CE+DA & Full & $.696 \pm .033$ & $.643 \pm .056$ \\
    SCL & Full & $.700 \pm .036$ & $.652 \pm .053$ \\ 
    DASCL & Full & $.709 \pm .021$ & $.640 \pm .041$ \\
    DASCL+DA & Full & $.718 \pm .031$ & $.628 \pm .060$ \\
    \hline 
    \end{tabular}
    \normalsize
    \caption{Precision and recall over the economic media test set \citep{barbera_boydstun_linn_mcmahon_nagler_2021} when using 100, 1000, and all labeled examples from the training set for fine-tuning. Except for the logistic regression model, results are averages over 10 random seeds with standard deviations reported.}  
    \label{tab:econmedia_results_test_precision_recall}
\end{table}

\subsubsection{Results over the Validation Set}
Table \ref{tab:econmedia_results_validation} reports the accuracy, precision, recall, and average precision over the validation set for the economic media data. The validation set was used for model selection, so the reported results over the validation set are from the model with the highest accuracy achieved on the validation set across the $5$ epochs.  

\begin{table*}
    \centering
    \begin{tabular}{ccccc}
    \hline 
    \textbf{Loss} & \textbf{Accuracy} & \textbf{Precision} & \textbf{Recall} & \textbf{Avg. Precision} \\ 
    \hline 
    CE & $.723 \pm .004$ & $.644 \pm .022$ & $.438 \pm .054$ & $.605 \pm .007$  \\
    CE w/ DA & $.726 \pm .004$ & $.662 \pm .035$ & $.423 \pm .059$ & $.607 \pm .007$ \\
    CE+SCL & $.727 \pm .003$ & $.659 \pm .037$ & $.438 \pm .065$ & $.611 \pm .006$ \\ 
    CE+DASCL & $.724 \pm .006$ & $.655 \pm .020$ & $.416 \pm .037$ & $.610 \pm .007$ \\
    CE+DASCL w/ DA & $.724 \pm .004$ & $.662 \pm .031$ & $.406 \pm .051$ & $.609 \pm .006$  \\
    \hline 
    \end{tabular}
    \caption{Accuracy, precision, recall, and average precision over the validation set for economic media \citep{barbera_boydstun_linn_mcmahon_nagler_2021}. Results are averages over 10 random seeds with standard deviations reported.}  
    \label{tab:econmedia_results_validation}
\end{table*}

\subsubsection{t-SNE Plots of the Learned Classifier Token Embeddings for the Test Set}
\label{sec:tsne_plots_econmedia}
\begin{figure*}
\centering
\begin{subfigure}{.45\textwidth}
  \centering
  \includegraphics[width=\linewidth]{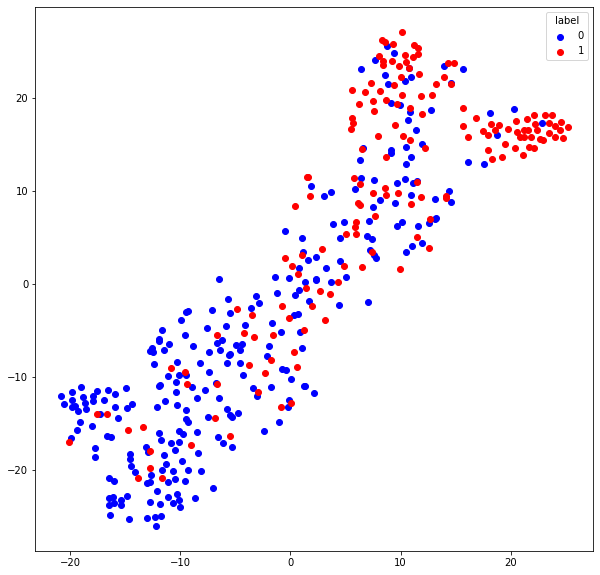}
  \caption{CE}
\end{subfigure}
\begin{subfigure}{.45\textwidth}
  \centering
  \includegraphics[width=\linewidth]{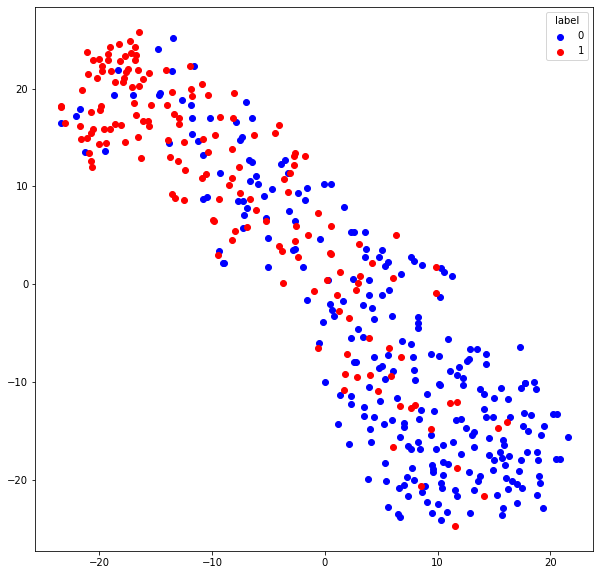}
  \caption{CE+SCL}
\end{subfigure}
\begin{subfigure}{.45\textwidth}
  \centering
  \includegraphics[width=\linewidth]{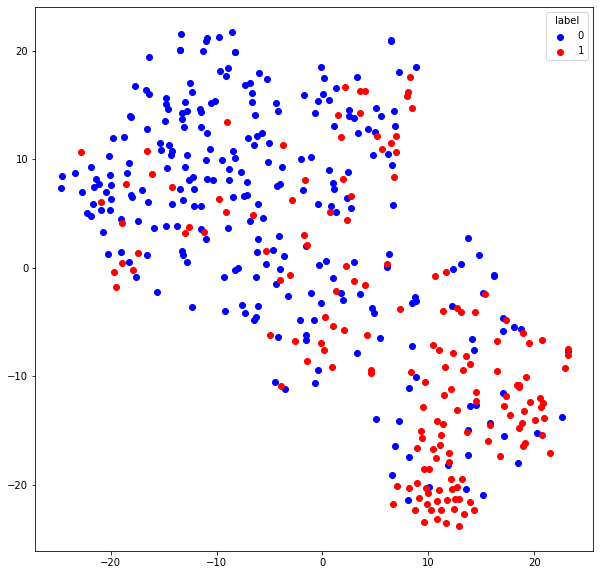}
  \caption{CE+DASCL}
\end{subfigure}
\caption{t-SNE plots of the classifier token embeddings on the \textit{New York Times} articles about the economy test set fine-tuned using the full training set. The loss configuration is noted below each plot. Blue are negative examples and red are positive examples.}
\label{fig:econmedia_tsne}
\end{figure*}

We use t-SNE plots to visualize the learned classifier token embeddings, ``$<$s$>$'', over the \textit{New York Times} articles about the economy test set when using the cross-entropy objective alone, using the cross-entropy objective with the supervised contrastive learning (SCL) objective \citep{gunel2020supervisedcontrastive}, and using the cross-entropy objective with the dictionary-assisted supervised contrastive learning (DASCL) objective. These plots are in Figure \ref{fig:econmedia_tsne}. We see that DASCL pushes embeddings of different classes farther apart compared to using cross-entropy alone or using cross-entropy with SCL.  

\subsection{Additional Information for the AbusEval Experiments}
\label{sec:abuseval_additional_info}
\subsubsection{Data Description: Few-Shot Training Set, Validation Set, and Test Set}
The data was downloaded from \url{https://github.com/tommasoc80/AbuseEval}. Because there was no validation set, we created our own validation set by sampling 15\% of the training set. 399 of the 1,986 tweets in the validation set are labeled abusive. For our few-shot training sets, when $N=100$, 17 tweets are labeled abusive. When $N=1000$, 210 tweets are labeled abusive. Our validation set and few-shot learning sets were sampled with no consideration to the label distributions of the original training or validation sets. 

\subsubsection{Text Preprocessing Steps}
We preprocessed the text of the tweets in the following manner: we removed all HTML tags, removed all URLs (even the anonymized URLs), removed the anonymized @ tags, removed the retweet (``RT'') tags, and removed all ``\&amp'' tags. The text is tokenized using the WordPiece tokenizer \citep{devlin-etal-2019-bert}. 

\subsubsection{Dictionaries Used During Keyword Simplification} 
We used two sets of dictionaries during keyword simplification. For the first dictionary, we used Lexicoder. For a description of the Lexicoder dictionary, see Section \ref{sec:econmedia_dictionaries}. We used the same token-replacements as described in Section \ref{sec:econmedia_dictionaries}. 

The second dictionary used was a dictionary of ``us'' and ``them'' pronouns. These pronouns are intended to capture directed or indirected abuse. The ``us'' pronouns are ``we're'', ``we'll'', ``we'd'', ``we've'', ``we'', ``me'', ``us'', ``our'', ``ours'', and ``let's''. The ``them'' pronouns are ``you're'', ``you've'', ``you'll'', ``you'd'', ``yours'', ``your'', ``you'', ``theirs'', ``their'', ``they're'', ``they'', ``them'', ``people'', ``men'', ``women'', ``man'', ``woman'', ``mob'', ``y'all'', and ``rest.'' This dictionary is loosely based on suggested words found in \citet{alorainy_othering}. 

\subsubsection{Number of Parameters and Runtime} 
\label{sec:abuseval_numparams_runtime}
This experiment uses the \textsc{BERT$_{\text{base-uncased}}$} pretrained language model, which contains 110 million parameters \citep{liu2019roberta}. When using DASCL, we also had an additional temperature parameter, $\tau$, that was directly optimized. With the hyperparameters described in Section \ref{sec:abuseval_hyperparam_configs} and using an NVIDIA V100 GPU, it took approximately 2.6 seconds to train over 40 batches using cross-entropy (CE) alone and 4.9 seconds to train over 40 batches using CE+DASCL. 

\subsubsection{Hyperparameter Selection and Configuration Details}
\label{sec:abuseval_hyperparam_configs}
We selected hyperparameters using the validation set. We searched over the learning rate and the temperature initialization; again, we used $\lambda = 0.9$ for all loss configurations involving contrastive learning. In line with \citet{caselli-etal-2021-hatebert}, we used a batch size of 32. We fine-tuned \textsc{BERT$_{\text{base-uncased}}$} and HateBERT for 5 epochs. 

For the learning rate, we searched over $\{1 \times 10^{-6}, 2 \times 10^{-6}, 3 \times 10^{-6}, 4 \times 10^{-6}, 5 \times 10^{-6}, 1 \times 10^{-5}, 2 \times 10^{-5} \}$; for the temperature, $\tau$, initialization, we searched over $\{0.07,0.3\}$. We fine-tuned the model and selected the model from the epoch with the highest F1 over the positive class. We repeated this with three random seeds, and selected the hyperparameter configuration with the highest average F1 over the positive class. We used the F1 score over the positive class as the criterion because it is one of the metrics of interest in \citet{caselli-etal-2021-hatebert}. The final learning rate across all loss configurations was $2 \times 10^{-6}$. The final $\tau$ initialization for both SCL and DASCL loss configurations was $0.3$. We note that our hyperparameter search yielded a different set of hyperparameters from \citet{caselli-etal-2021-hatebert}. We used these same hyperparameters when we limited the training set to 100 and 1,000 labeled examples. 

\subsubsection{Model Evaluation Details} 
\label{sec:abuseval_model_eval_details}
During fine-tuning, the model from the epoch with the highest F1 over the validation set was chosen as the final model for each random seed. We report macro F1 and F1, the main metrics of interest with this dataset, and average precision. For a definition of average precision, see Section \ref{sec:sst2_model_eval_details}. 

\subsubsection{Results over the Validation Set} 
Table \ref{tab:abuseval_results_validation} reports the macro F1, F1, and average precision over the validation set for AbusEval. The validation set was used for model selection, so the reported results over the validation set are from the model with the highest F1 achieved on the validation set across the $5$ epochs.  

\begin{table}
    \small 
    \centering
    \addtolength{\tabcolsep}{-2.5pt}  
    \begin{tabular}{cccc}
    \hline 
    \textbf{Model} & \textbf{Macro F1} & \textbf{F1, Pos} & \textbf{Avg. Precision}\\
    \hline 
    BERT & $.756 \pm .006$ & $.632 \pm .008$ & $.681 \pm .014$  \\ 
    HateBERT & $.754 \pm .009$ & $.635 \pm .008$ & $.708 \pm .005$  \\
    \begin{tabular}[c]{@{}c@{}}BERT+\\DASCL\end{tabular} & $.759 \pm .007$ & $.639 \pm .007$ & $.683 \pm .012$ \\
    \begin{tabular}[c]{@{}c@{}}HateBERT+\\DASCL\end{tabular} & $.755 \pm .005$ & $.635 \pm .005$ & $.706 \pm .006$ \\
    \hline 
    \end{tabular}
    \caption{The macro F1, F1, and average precision over the AbusEval validation set \citep{caselli-etal-2020-feel}. Results are averages over 10 random seeds with standard deviations reported.}
    \label{tab:abuseval_results_validation}
\end{table}

\subsubsection{t-SNE Plots of the Learned Classifier Token Embeddings for the Test Set}
\label{sec:tsne_plots_abuseval}
\begin{figure*}
\centering
\begin{subfigure}{.45\textwidth}
  \centering
  \includegraphics[width=\linewidth]{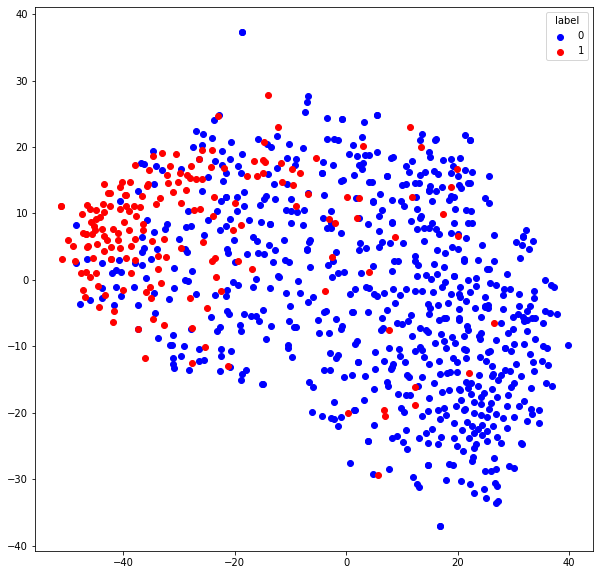}
  \caption{BERT}
\end{subfigure}
\begin{subfigure}{.45\textwidth}
  \centering
  \includegraphics[width=\linewidth]{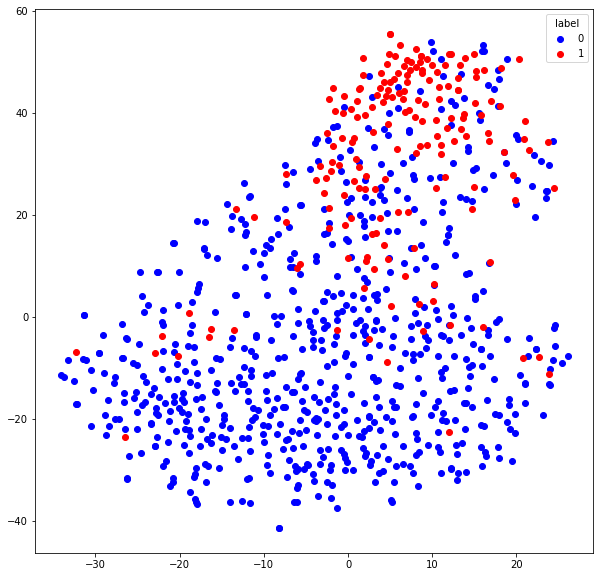}
  \caption{BERT with DASCL}
\end{subfigure}
\caption{t-SNE plots of the classifier token embeddings on the AbusEval test set fine-tuned using the full training set. The model is noted below each plot. Blue are examples of non-abusive tweets and red are examples of abusive tweets.}
\label{fig:abuseval_tsne}
\end{figure*}

We use t-SNE plots to visualize the learned classifier token embeddings, ``$<$s$>$'', over the AbusEval test set when using BERT and when using BERT with DASCL. These plots are in Figure \ref{fig:abuseval_tsne}. We see that using DASCL with BERT pushes embeddings of different classes farther apart compared to using BERT alone.  

\end{document}